\documentclass[10pt,twocolumn,letterpaper]{article}

\usepackage{wacv}              

\newcommand{\red}[1]{{\color{black}#1}}

\usepackage{pgfplots}
\pgfplotsset{compat=1.18}

\usepackage{amsmath,amssymb,amsfonts,mathtools}

\usepackage{amsmath}
\usepackage{amsthm}

\newtheorem{proposition}{Proposition}

\usepackage{graphicx}
\usepackage{booktabs}
\usepackage{multirow}
\usepackage{colortbl}
\usepackage{makecell}
\usepackage{hhline}
\usepackage{wrapfig}
\usepackage{adjustbox}
\usepackage{float}
\usepackage{multicol}

\usepackage{algorithm}
\usepackage{algpseudocode}

\usepackage[dvipsnames]{xcolor}
\definecolor{mydarkgreen}{rgb}{0.0, 0.5, 0.0}
\definecolor{mydarkred}{rgb}{0.6, 0.0, 0.0}

\usepackage{xfrac}
\usepackage{nicefrac}
\usepackage{url}
\usepackage{comment}
\usepackage{needspace}

\newcommand{\mypar}[1]{\vspace{4pt}\noindent\textbf{#1}~}

\newcommand{\Real}{\mathbb{R}}

\newcommand{\tr}[1]{#1^{\top}}

\newcommand{\One}{\mathbf{1}}

\newcommand{\mr}[1]{\mathrm{#1}}

\usepackage{scalerel}

\newcommand\wt[1]{\hstretch{1.25}{\tilde{\hstretch{.8}{#1}}}}

\newcommand{\rcol}{\rowcolor{Dandelion!12}}
\cellcolor{Dandelion!12}
\newcommand{\improve}[1]{\textcolor{BrickRed}{\,\scriptsize+#1}}
\newcommand{\decrease}[1]{\textcolor{RoyalBlue}{\,\scriptsize-#1}}

\usepackage{glossaries}

\newacronym{cnn}{CNN}{Convolutional Neural Network}
\newacronym{densenet}{DenseNet}{Dense Convolutional Network}
\newacronym{vit}{ViT}{Vision Transformer}
\newacronym{nlp}{NLP}{Natural Language Processing}
\newacronym{ssm}{SSM}{State Space Model}
\newacronym{deit}{DeiT}{Data-efficient image Transformers}
\newacronym{ss2d}{SS2D}{2D Selective Scan}
\newacronym{dpct}{DPCT}{Dynamic Patch Connectivity Traversal}
\newacronym{csm}{CSM}{Corss-Scan Module}
\newacronym{iam}{IAM}{Isometric Adjustment Module}
\newacronym{ema}{EMA}{Exponential Moving Average}
\newacronym{knn}{KNN}{K-Nearest Neighbors}
\newacronym{miou}{mIoU}{mean Intersection over Union}
\newacronym{vim}{ViM}{Vision Mamba}
\newacronym{msv}{MSVMamba}{Multi-Scale VMamba}
\newacronym{zoh}{ZOH}{zero-order hold}
\newacronym{rfn}{RFN}{Rotational Feature Normalizer}
\newacronym{sts}{STS}{Spectral Traversal Scan}
\newacronym{stm}{STM}{Spectral Traversal Merge}
\newacronym{erf}{ERF}{Effective Receptive Field}
\newacronym{lsa}{LiSA}{Linear Stochastic Attention}
\newacronym{la}{linear-attention}{}
\newacronym{sa}{self-attention}{}
\newacronym{soa}{softmax-attention}{}

\definecolor{wacvblue}{rgb}{0.21,0.49,0.74}
\usepackage[pagebackref,breaklinks,colorlinks,allcolors=wacvblue]{hyperref}

\def\wacvPaperID{1115} 
\def\confName{WACV}
\def\confYear{2027}

\title{LiSA: Linear Stochastic Attention for Long-Range Visual Modeling}

\author{
Sahar Dastani$^{1,2}$,
Fereshteh Shakeri$^{1}$,
Samuel Barbeau$^{1}$,
Kunal Mahatha$^{1}$,\\
Ismail Ben Ayed$^{1}$,
Herve Lombaert$^{2,3}$,
Christian Desrosiers$^{1}$\\[0.5em]
\mbox{\fontsize{9}{9}\selectfont
$^{1}$LIVIA, ÉTS Montreal, Canada
\quad
$^{2}$Mila -- Quebec AI Institute, Canada
\quad
$^{3}$Polytechnique Montreal, Canada
}
}

\begin{document}
\maketitle
\begingroup
\renewcommand{\thefootnote}{}
\footnotetext{\scriptsize
\textcopyright~2027 IEEE.
Personal use of this material is permitted.
Permission from IEEE must be obtained for all other uses,
in any current or future media, including
reprinting/republishing this material for advertising
or promotional purposes, creating new collective works,
for resale or redistribution to servers or lists,
or reuse of any copyrighted component of this work
in other works.
}
\endgroup
\begin{abstract}
    Self-attention has become the standard for modern sequence modeling, enabling Transformers to model long-range dependencies across a wide range of modalities. Despite these advances, its quadratic cost with respect to sequence length remains a significant computational bottleneck. Recent work addresses this limitation by introducing linear variants of self-attention, namely with kernel approximations of the softmax operator. However, their first-order aggregation restricts expressivity and struggles to capture higher-order structure. We propose Linear Stochastic Attention (LiSA), a linear attention plug-in module that enables higher-order token interactions via a stochastic diffusion process based on truncated random walks. LiSA incorporates adaptive locality, favoring semantically similar neighbors to preserve fine-grained structures, while maintaining stable and efficient training. We conduct extensive experiments on classification, semantic segmentation, and object detection tasks and show improved results over state-of-the-art methods. The code is available \href{https://github.com/Sahardastani/lisa.git}{here}.
\end{abstract}
 
\vspace{-6mm}
\section{Introduction}
\label{sec:intro}

Transformers \cite{attention} have become a central paradigm in visual representation learning, with \gls{sa} \cite{lin2017structuredselfattentivesentenceembedding} serving as a key mechanism for capturing long-range dependencies. Since the advent of \glspl{vit}~\cite{vit,deit}, \gls{sa} has delivered strong performance across image classification \cite{deit, pvt}, object detection \cite{vitdet, detr}, segmentation \cite{maskformer, segformer}, and multi-modal tasks~\cite{clip}, highlighting the value of global context in modern vision pipelines. However, \gls{sa} scales quadratically with token length, which limits applicability to high-resolution images.

A number of recent works seek to reduce the quadratic complexity of standard \glspl{vit}. One common approach restricts attention to a limited set of tokens, either through fixed-sized attention windows \cite{SwinTransformer, quadtree}, introducing sparsity \cite{pvt, dat} or representing the attention queries and keys with low rank matrices \cite{linformer}. By reducing the number of pairwise token interactions, these methods lower computational cost, but often at the expense of global context modeling.

\begin{figure}[tb]
  \centering
    \includegraphics[width=\columnwidth]{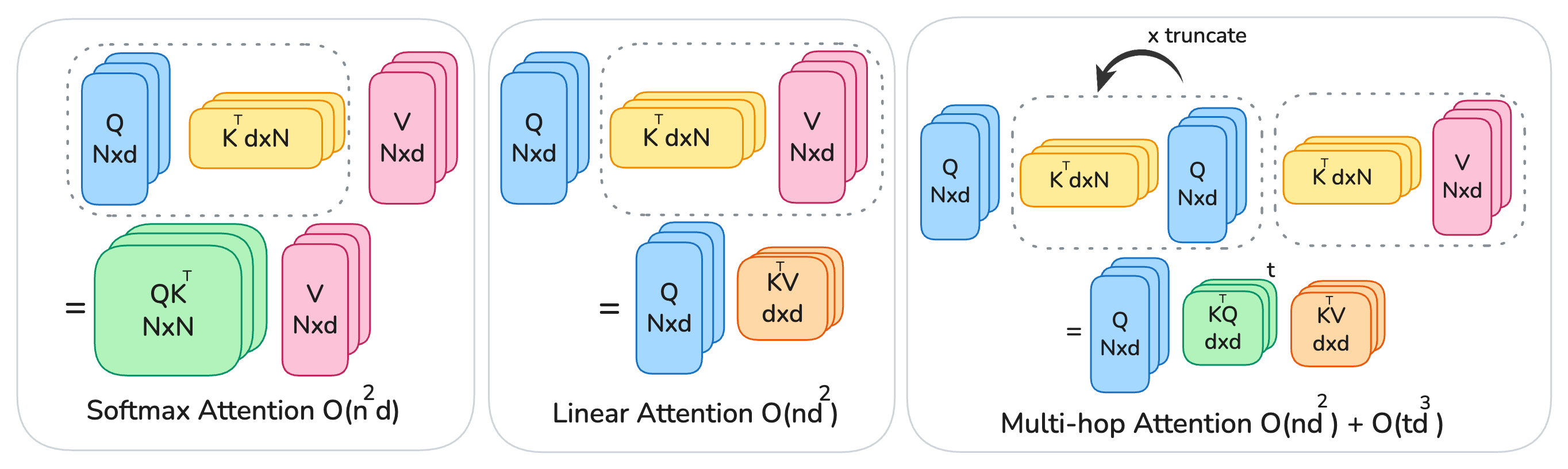}  
    \vspace{-7mm}
    \caption{\textbf{Softmax vs. Linear vs. Higher-order attention.}
      Given $Q,K,V\!\in\!\mathbb{R}^{n\times d}$:
      \gls{soa} forms $QK^\top$ then applies to $V$.
      Linear-attention computes $K^\top V$ then applies to $Q$. \red{Higher-order attention (ours) uses a truncation length $T$, retaining multiple diffusion terms while remaining linear in $n$ and expanding the effective receptive field.}
      }
  \label{fig:example}
  \vspace{-7mm}
\end{figure}

Other approaches linearize \gls{sa} using kernel-based approximations \cite{tranformers_rnns, hydra, efficientattn, cosformer}, replacing the softmax with a positive feature map $\phi(\cdot)$. This enables an associative reordering of the $Q$, $K$, and $V$ computation that avoids explicit pairwise interactions and reduces the complexity to linear in the sequence length. These methods retain a global receptive field without relying on downsampling.

In practice, however, an accuracy gap between kernelized \gls{la} and standard \gls{sa} persists. Recent works have reduced this gap by linearizing dot-product attention \cite{efficientvit}, enriching the kernel with angular \cite{castling} or focused parameterizations, and adding depthwise-convolutional components \cite{flatten} or auxiliary softmax branches \cite{castling}. However, these variants are almost always built on the same core backbone, vanilla kernelized \gls{la} \cite{tranformers_rnns}. Our analysis points to a structural bottleneck in this baseline: its single-pass bilinear aggregation.

Vanilla \gls{la} mixes values for each query once through direct query-key affinities. With the affinity matrix defined as $A\,{=}\,\phi(Q)\,\phi(K)^\top$, the output $Y\,{=}\,A V$ can be interpreted as a first-order message-passing step along the edges encoded by $A$. Crucially, indirect dependencies (i.e., using intermediate tokens) would appear only in higher powers (e.g., $A^2 V, A^3 V$), but vanilla \gls{la} never materializes higher-order terms. As a result, when two regions are semantically related yet have weak first-order similarity (e.g., appearance changes due to lighting, occlusion, or viewpoint), their connection is under-weighted and the signal is diluted by many competing matches. In contrast, \gls{soa} introduces competitive normalization that can concentrate probability mass on high-scoring connections, which help preserve long-range links.

Our approach targets this limitation and proposes a novel plug-in \emph{\gls{lsa}} module, which achieves efficient higher-order information propagation without sacrificing stability or locality. \gls{lsa} models attention as a linear stochastic process, where image patches are represented as graph nodes and edges encode transition probabilities among them during a truncated random walk. This formulation allows information to traverse intermediate tokens, enabling transitive reasoning. The random walk is truncated to a small number of steps to maintain linear complexity and numerical stability, providing a controllable balance between expressiveness and efficiency.

Overall, \gls{lsa} serves as a stronger drop-in module: it integrates cleanly with standard \gls{la} refinements and often amplifies their benefits, delivering higher accuracy than vanilla \gls{la} while preserving linear-time complexity and training stability. Our contributions are summarized as follows:

\begin{itemize}
\renewcommand{\labelitemi}{\textbullet}
    \item We introduce \gls{lsa}, a plug-in \gls{la} module that enables efficient higher-order token interactions.

    \item We formulate attention as a truncated random walk, enabling transitive information flow while preserving linear-time complexity and stable training.

    \item We evaluate our method across classification, segmentation, and detection, consistently outperforming both \gls{sa} and \gls{la} baselines.
\end{itemize}

\vspace{-3mm}
\section{Related Work}
\label{sec:related}

\subsection{Efficient Vision Transformer}

\red{Despite their success in computer vision, attention-based models are computationally expensive due to the quadratic scaling of \gls{sa} with the number of tokens. To reduce this cost, \gls{vit}, VOLO, and T2T-ViT \cite{vit,volo,t2tvit} use patchification or progressive tokenization, with similar approaches applied to downstream tasks \cite{vitdet}. Other methods reduce attention costs by restricting token interactions: PVT/PVTv2 \cite{pvt,pvtv2} reduce key/value resolution, Swin \cite{SwinTransformer} uses local windows, NAT \cite{NAT} uses neighborhood attention, and BiFormer \cite{biformer} adopts bi-level routing. However, the cost of \gls{soa} remains a challenge for high-resolution images.}

\subsection{Linear Attention}

\red{An alternative to quadratic self-attention is Linear attention (\gls{la})~\cite{tranformers_rnns}, which reduces the quadratic cost of self-attention by factorizing query--key interactions and computing $K^\top V$, reducing complexity from $\mathcal{O}(n^2d)$ to $\mathcal{O}(nd^2)$ while preserving the global context.

To improve the efficiency and expressiveness of \gls{la}, Performer uses random features~\cite{performer}, CosFormer uses cosine re-weighting~\cite{cosformer}, SOFT adopts low-rank approximations~\cite{nystromformer,soft_free}, and Efficient Attention separate query/key normalization~\cite{efficientattn}. FLatten token selectivity~\cite{flatten}, and Hydra and Castling-ViT explore alternative similarity functions~\cite{hydra,castling}. More recently, MILA adopts Mamba-inspired designs~\cite{demystify_mamba}, InLine addresses non-injectivity and locality~\cite{inline}, RALA mitigates the low-rank bottleneck~\cite{rala}, and PolaFormer models polarity-aware interactions~\cite{polaformer}. Despite these advances, most \gls{la} methods primarily model direct token interactions. Our \gls{lsa} captures higher-order, transitive interactions through truncated random walks while preserving linear complexity and achieving competitive accuracy compared to \gls{soa}.}

\vspace{-5pt}
\subsection{Higher-Order Attention}

\red{Several works extend attention beyond direct interactions through diffusion or graph filtering. MAGNA~\cite{magna} diffuses attention over multiple hops in graph neural networks, while Diffuser~\cite{diffuser} extends sparse Transformer attention with multi-hop diffusion. From a graph-signal-processing perspective, GFSA~\cite{gfsa} interprets self-attention as a first-order graph filter and generalizes it through polynomial filtering. More recently, AGF~\cite{agf} extends this perspective to linear Transformers by learning graph filters directly in the singular-value domain while retaining linear complexity. Attention Chains~\cite{attn_chains} instead interprets existing softmax attention matrices as Markov chains, using repeated transitions to characterize indirect attention and stationary token importance.

Our \gls{lsa} also models higher-order interactions, but differs in how they are computed. Unlike MAGNA and Diffuser, which diffuse attention over graph or sparse-attention structures, \gls{lsa} constructs a stochastic transition matrix directly from kernelized query--key affinities. While this can also be viewed as polynomial graph filtering, unlike AGF's spectral formulation, \gls{lsa} uses non-negative, normalized step weights to control multi-hop propagation. This allows information to flow through intermediate tokens while maintaining a probabilistic interpretation, with a local attention branch to preserve spatial details. Unlike Attention Chains, \gls{lsa} learns these interactions end-to-end to produce layer representations. Its low-rank factorization computes higher-order terms in the $d_h\times d_h$ space, preserving linear complexity in the number of tokens.}

\vspace{-2mm}
\section{Method}
\label{sec:method}

In Section~\ref{sec:preli} we give a brief definition of the \gls{sa} mechanism and the linear version. We then proceed with the description and formulation of our proposed \gls{lsa} module in Section~\ref{sec:lsa}.

\begin{figure*}[!t]
\centering
\includegraphics[width=\textwidth]{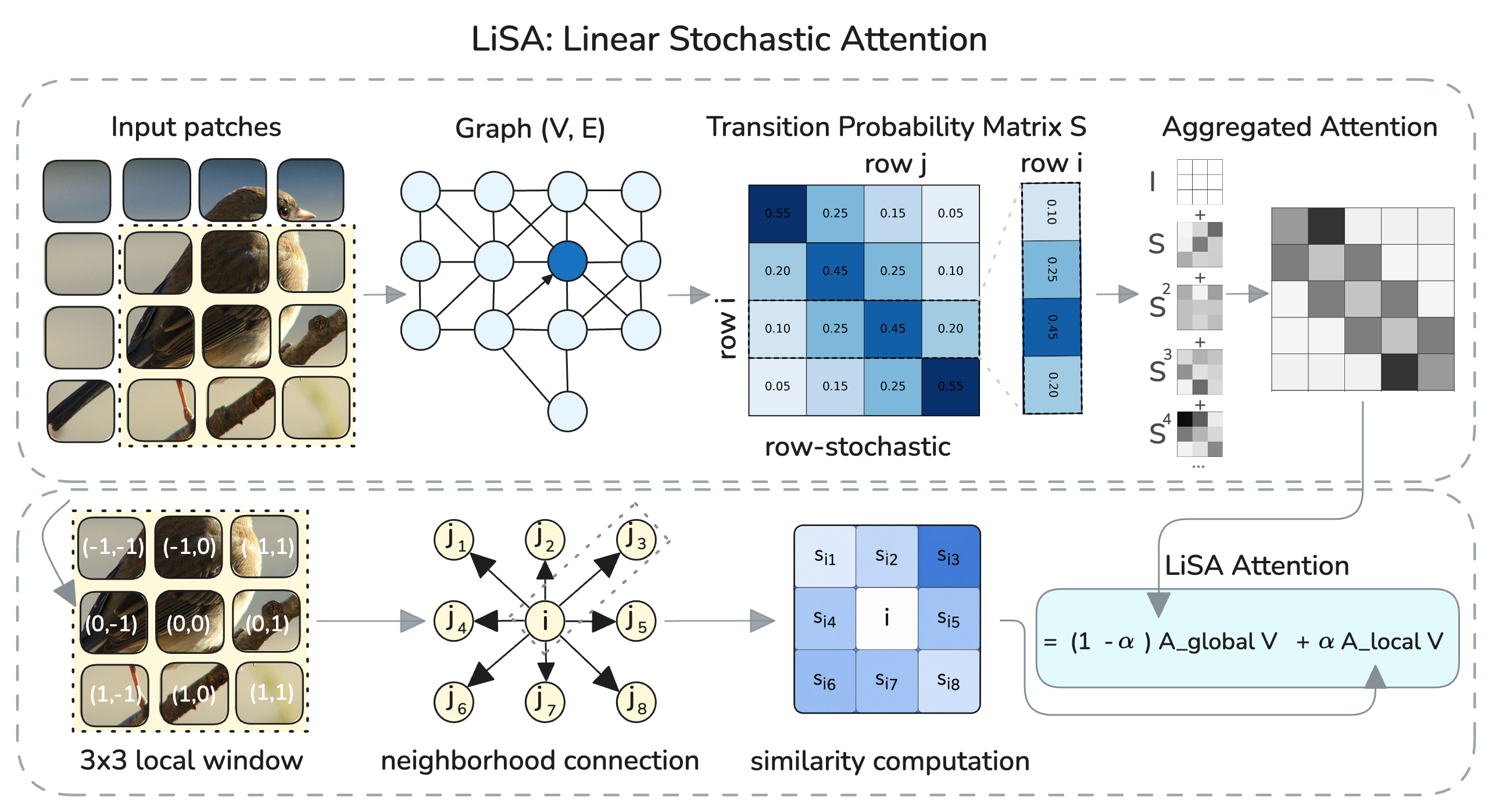}
\caption{\red{Overview of \gls{lsa}. We patchify the input image into tokens and treat them as nodes in a graph \(G=(V,E)\), where edge weights define a row-stochastic transition matrix \(S\). Aggregating the truncated terms on \(S\) yields a global attention map that captures transitive (higher-order) dependencies. In parallel, we compute content-based similarity within each token’s local neighborhood to form a local attention map. The layer output is a convex combination of the two, blending sharp local context with long-range structure.}}
\label{sts}
\vspace{-10pt}
\end{figure*}

\subsection{Preliminaries}
\label{sec:preli}

\mypar{Self-attention (SA).} We start by presenting the \gls{sa} model of transformers. Given a set of $n$ tokens $X \in \Real^{n\times d}$, each represented by a $d$-dimensional vector, \gls{sa} maps $X$ to a set of queries $Q\,{=}\,XW_Q \in \Real^{n \times d_h}$, keys $K\,{=}\,XW_K \in \Real^{n \times d_h}$ and values $V\,{=}\,XW_V \in \Real^{n \times d}$, where $W_Q$, $W_K$ and $W_V$ are parameter matrices learned in training. The output of the \gls{sa} layer is obtained as:
\begin{equation}\label{eq:sa}
Y \, = \, \mathrm{softmax}\Big(\frac{Q\tr{K}}{\sqrt{d}}\Big)V
\end{equation}
The output can also be expressed for a specific token $i$ as:
\begin{equation}
y_i \;=\; \sum_{j=1}^N a_{ij}\, v_j,\quad \text{where} \quad
a_{ij} \;=\; \frac{\exp\!\left(\nicefrac{q_i^\top k_j}{\sqrt{d}}\right)}{\sum_{j'=1}^N \exp\!\left(\nicefrac{q_i^\top k_{j'}}{\sqrt{d}}\right)},
\end{equation}
and $y_i, v_j \in \Real^{d}$. Self-attention requires computing the attention score $a_{ij}$ between each pair of tokens before the softmax, leading to $\mathcal{O}(n^2)$ time and memory complexity. This quadratic scaling makes high-resolution inputs and long sequences prohibitively expensive.

In this work, we seek an alternative to \gls{sa} with the following properties: (1) \textbf{Scalability}: The complexity of the model in terms of memory and computation should increase \emph{linearly} with respect to the number of tokens, during training as well as during inference. (2) \textbf{Expressiveness}: \red{The model should also have expressiveness similar to that of \gls{sa} while maintaining a comparable parameter budget.} In particular, it should capture \emph{global} and \emph{non-linear} relations between input tokens.

\mypar{Linear Attention (LA).} \Gls{la}~\cite{tranformers_rnns} offers an effective alternative to \gls{soa} by reducing the complexity from $\mathcal{O}(n^2)$ to $\mathcal{O}(n)$ in the sequence length $N$. The key idea is to replace the softmax similarity with a positive feature map $\phi(\cdot)$ so that the pairwise similarity can be written in a kernelized form $\phi(Q)\phi(K)^{\!\top}$. With this factorization, the output for token $i$ becomes:
\begin{equation}
y_i = \frac{\sum_{j=1}^{N} \Big(\phi(q_i)\, \phi(k_j)^{\!\top}\Big) v_j}{\sum_{j=1}^{N} \Big(\phi(q_i)\, \phi(k_j)^{\!\top}\Big)}
 = \frac{\phi(q_i) \Big(\!\sum_{j=1}^{N} \phi(k_j)^{\!\top} v_j\!\Big)}
{\phi(q_i) \Big(\!\sum_{j=1}^{N} \phi(k_j)^{\!\top}\!\Big)}
\end{equation}
thereby avoiding the $n \times n$ attention matrix and yielding linear-time computation.

Despite their efficiency, current \gls{la} approaches primarily perform a single-pass aggregation: $Y\,{=}\,\phi(Q)\big(\phi(K)^{\!\top} V\big)$ mixes each query with values exactly once via first-order (one-hop) affinities. This design fails to materialize transitive dependencies involving intermediate tokens, which would arise from higher powers of $A\,{=}\,\phi(Q)\phi(K)^\top$.

\subsection{Linear Stochastic Attention (LiSA)}
\label{sec:lsa}

Our \gls{lsa} approach is based on discrete random walks. This stochastic process models the probability of moving among the nodes $\mathcal{V}$ of a graph $\mathcal{G} = (\mathcal{V},\mathcal{E})$ with a time-invariant transition probability matrix $S \in \Real^{|\mathcal{V}| \times |\mathcal{V}|}$ where each $s_{ij}$, for $(i,j)\in\mathcal{E}$ defines the probability of moving from node $i$ to node $j$ at any given time. Matrix $S$ is row-stochastic as $\sum_j s_{ij}=1$, $\forall i\in\mathcal{V}$. 

At each step, the random walk can either continue by moving to another node with probability $0 \!\leq\! \beta \!<\! 1$, in which case the next node is selected based on the transition probability matrix $S$, or  stop with probability $(1-\beta)$. \red{Denote by $p^{(T)}_{ij}$ the probability of ending the walk at node $j$ when starting at node $i$ under a truncation length $T$. Here, $T$ denotes the truncation length, i.e., the number of terms retained in the diffusion expansion, corresponding to powers $S^0,\ldots,S^{T-1}$.} The random walk probabilities can be expressed as a geometric series:
\red{\begin{equation}\label{eq:geometric}
P^{(T)} \, = \, \left(\frac{1\!-\!\beta}{1\!-\!\beta^{T}}\right)\sum_{k=0}^{T-1} \beta^k S^k.
\end{equation}}

\vspace{-7pt}
\begin{proposition}
For walks of infinite length, probabilities can be computed analytically as:
\begin{equation}\label{eq:inverse}
P^{(\infty)} \, = \, (1\!-\!\beta)\big(I - \beta S\big)^{-1}.
\end{equation}
The proof is provided in the supplementary material.
\end{proposition}

To model global relationships between tokens, we follow the \gls{sa} approach based on matrices representing the $K$, $Q$ and $V$ of tokens. However, we consider a more general formulation $Q=\phi(XW_Q) \in \Real^{n \times d_h}$, $K=\phi(XW_K) \in \Real^{n \times d_h}$, $V=XW_V \in \Real^{n \times d}$, where $\phi$ is a point-wise activation function.

Considering tokens as nodes in the graph, the transition probability from a token $i$ to a token $j$ is then computed as:
\begin{equation}
s_{ij} \, = \, \frac{q_i \tr{k}_j}{\sum_{j'} q_i \tr{k}_{j'}}. 
\end{equation}
This can be expressed in matrix form as: $S \, = \, (D^{-1}Q)\tr{K} \, = \, \wt{Q}\tr{K}$
with $D = \mr{Diag}(Q(\tr{K}\One))$. To avoid division by zero, we add a small constant $\epsilon=10^{-6}$ to the normalization denominator. Intuitively, the transition probability $s_{ij}$ encodes information similar to the attention score $a_{ij}$ of \gls{sa}, where a higher value means a stronger relationship between tokens $i$ and $j$.

To obtain the transformed features of a token $i$, we imagine a generative process where a random walk is initiated at node $i$ and, once the walk stops at a node $j$, the value vector $v_j$ is generated. For a potentially infinite random walk with continuation probability $\beta$, the expected features are then given by:
\red{\begin{equation}\label{eq:token_feat}
Y_{\text{global}}
= A_{\text{global}} V = (1\!-\!\beta)
\left(\overbrace{\sum_{k=0}^\infty \beta^k (\wt{Q}\tr{K})^k}^{\big(I - \beta \wt{Q}\tr{K}\big)^{-1}}\right)
V.
\end{equation}}

The formulation in Eq. (\ref{eq:token_feat}) satisfies the desired property of expressiveness, as the function models an infinite-order relation between the input $X$ and output $Y$ with the same number of trained parameters as \gls{sa}. Moreover, it can be computed effectively, with a complexity linear in the number of tokens $n$, using the property that $S=\wt{Q}\tr{K}$ is low rank. Nevertheless, matrix inversion, and in particular computing the derivative of this operation during back-propagation, remains expensive. In the next section, we present a more efficient solution based on truncated random walks.

\mypar{Truncated Random Walk Approximation.} \red{To avoid matrix inversions, we retain the first $T$ terms of the random-walk expansion.} In this case, the computation of output features becomes:
\red{\begin{equation}\label{eq:truncated}
\begin{aligned}
Y_{\text{global}} & \,=\, \left(\frac{1\!-\!\beta}{1\!-\!\beta^{T}}\right)\bigg(\sum_{k=0}^{T-1} \beta^k S^k\bigg)V\\
 & \,=\, \left(\frac{1\!-\!\beta}{1\!-\!\beta^{T}}\right)\bigg(I + \sum_{k=1}^{T-1} \beta^k\wt{Q}(\tr{K}\wt{Q})^{k-1}\tr{K}\bigg)V\\
& \,=\, \left(\frac{1\!-\!\beta}{1\!-\!\beta^{T}}\right)\Big(V + \beta\wt{Q}\bigg(\underbrace{\sum_{k=0}^{T-2} \beta^k(\tr{K}\wt{Q})^{k}}_{d_h \times d_h}\bigg)\tr{K}V\Big)
\end{aligned}
\end{equation}}

\noindent This gives rise to the efficient computational process in Algorithm \ref{alg:truncate}. Each loop of this algorithm multiples two $d_h\,{\times}\,d_h$ matrices, which has a complexity $\mathcal{O}(d_h^3)$. Therefore, the total complexity of the algorithm is $\mathcal{O}(T \cdot d_h^3) + \mathcal{O}(n \cdot d \cdot d_h)$. Due to the added $T$ factor, the truncation strategy has the important advantage of requiring only matrix multiplications that can be easily parallelized on GPU.

\noindent However, since this approach limits the random walk to a finite length, it can produce results that differ from those of the original infinite random walk model. Nevertheless, as shown in the following proposition, this discrepancy rapidly diminishes as the walk length increases.

\Needspace{8\baselineskip}
\begin{proposition}
\label{theorem2}
\red{Let $R_T$ be the truncation residual obtained after retaining the first $T$ terms in $P^{(\infty)}$. The $L_1$ norm of $R_T$ decays exponentially as the truncation depth $T$ increases.}
\red{\begin{equation}
R_T \, = \, (1\!-\!\beta)\left(\sum_{k=T}^{\infty}\!\!\beta^k S^k\right). 
\end{equation}}
The proof is provided in the supplementary material.
\end{proposition}

\mypar{Learnable Step Weights.} To further increase the expressiveness of the model, we allow the step weights in Eq.~(\ref{eq:truncated}) to be learnable and input-dependent. Specifically, we replace the fixed geometric weights $\left(\frac{1-\beta}{1-\beta^{T}}\right)\beta^k$ with non-negative weights $\beta_k$ predicted from the input and normalized across the $T$ retained terms. This yields:
\red{\begin{equation}
Y_{\text{global}}
=
\beta_0 V
+
\sum_{k=0}^{T-2}
\beta_{k+1}\,
\wt{Q}
(\tr{K}\wt{Q})^{k}
\tr{K}V,
\label{eq:learnable}
\end{equation}}
\red{where $\beta_k \geq 0$ and $\sum_{k=0}^{T-1}\beta_k = 1$. Further details on the generation of the step weights are provided in the supplementary material.}

\mypar{Link to softmax in SA.} \red{Softmax attention adds nonlinearity through the \emph{pointwise} exponential function: $\exp(x){=} \sum_{k=0}^{\infty}\! \frac{x^k}{k!}$. In contrast, the random-walk (RW) diffusion probabilities are defined by a matrix power series over the \emph{transition matrix} $S$: $P^{(\infty)} = (1{-}\beta)\sum_{k=0}^{\infty}\beta^k S^k$. Because each power $S^k$ captures $k$-step transitions, RW diffusion models higher-order, non-pointwise relationships across the graph.}

\mypar{Adaptive Local Attention.}

\red{In LiSA, increasing the truncation length $T$ retains higher-order powers of the row-stochastic transition operator, progressively extending information propagation across patches.} Since each step updates $\mathbf{x}_i$ as a weighted average of other tokens, $\|\mathbf{x}_i-\mathbf{x}_j\|$ shrinks over steps, making distinct regions (e.g., foreground and background) harder to separate (over-smoothing). We add a local branch that restricts mixing to a $3\times3$ neighborhood to preserve spatial detail, while the global branch provides long-range context. This module models local interactions within a small neighborhood window around each token. For each patch token, affinities are computed only among its immediate spatial neighbors, with higher transition probabilities assigned to semantically similar regions. Let the spatial position of token $i$ be 
$(h_i, w_i)\in\mathbb{Z}^2$ and define:
\begin{equation}
\begin{aligned}
\mathcal{N}(i)=\big\{\, j :\; 
& (h_j, w_j) = (h_i, w_i) + (\Delta_h,\Delta_w),\\
& (\Delta_h,\Delta_w)\in\{-1,0,1\}^2
\,\big\}.
\end{aligned}
\end{equation}
We then form a sparse local affinity matrix $A^{\text{local}}\in\mathbb{R}^{N\times N}$ whose non-zero entries connect a node $i$ only to its spatial neighbors:
\vspace{-9pt}
\begin{equation}
\begin{aligned}
A_{\text{local}} &= M \odot \exp\left(\frac{QK^{\!\top}}{\sqrt{d}}\right)
\end{aligned}
\end{equation}
where $M_{ij} = \mathbf{1}\{\,j\in\mathcal{N}(i)\,\}$. We row-normalize this affinity matrix to obtain the local attention transition. The neighborhood includes a self-transition through the $(0,0)$ offset, which helps stabilize local aggregation. Without this self-loop, each token would be forced to replace itself with a weighted average of its neighboring tokens. All entries outside $\mathcal{N}(i)$ are zero and are never explicitly computed, so the local branch has linear complexity in the number of tokens.

\begin{algorithm}[t]
\caption{Truncation-based computation}
\label{alg:truncate}
\begin{algorithmic}
\Require Truncation length $T$
\State $M \gets I$
\State $A \gets \beta\tr{K}\wt{Q}$
\State $Y \gets V$
\State $B \gets \tr{K}V$
\For{$k=1,\ldots,T\!-\!1$}
    \State $Y \gets Y + \beta\wt{Q}(MB)$
    \State $M \gets A \cdot M$
\EndFor
\State $Y \gets
\Big(\frac{1-\beta}{1-\beta^{T}}\Big)Y$
\State \Return $Y$
\end{algorithmic}
\end{algorithm}

\mypar{Mixture of Operators.} \red{We combine the outputs of the global and local branches, allowing the model to trade off long-range higher-order aggregation with local neighborhood refinement using a learnable gate:
\begin{equation}
Y=(1-\alpha) \underbrace{A_{\mathrm{global}} V}_{Y_\mathrm{global}} +\alpha \underbrace{A_{\mathrm{local}} V}_{Y_\mathrm{local}},
\label{eq:fused_output}
\end{equation}
where $\alpha\in[0,1]$ is a learnable parameter that balances global and local information.}

\section{Experiments}
\label{sec:exps}

We validate the effectiveness of our method on ImageNet-1K classification \cite{imagenet}, ADE20K semantic segmentation \cite{ade20k}, and COCO object detection \cite{coco}, and include extensive ablations to assess each design component.

\subsection{Image Classification} \label{sec:classification}

\mypar{Datasets.} We use the ImageNet-1K dataset \cite{imagenet} for image classification. This dataset comprises 1.28M training images and 50K validation images across 1,000 categories. 

\mypar{Settings.} We integrate our module into four Vision Transformer backbones, DeiT \cite{deit}, PVT \cite{pvt}, Swin Transformer \cite{SwinTransformer}, and BiFormer \cite{biformer} and report Top-1 accuracy on the validation split. For a fair comparison, we adopt the same training recipe as the baseline. All models are trained from scratch for 300 epochs with AdamW \cite{adamw}, a cosine-decayed learning rate and a 20-epoch linear warmup. We initialize the LR at $1\,{\times}\,10^{-3}$ and use a weight decay of 0.05. Data augmentation and regularization include RandAugment \cite{randaugment}, Mixup \cite{mixup}, CutMix \cite{cutmix}, and Random Erasing \cite{random_erasing}. The experiments were conducted on eight NVIDIA A100/H100 GPUs. \red{We replace every \gls{sa} layer with our \gls{lsa} module and use $T\,{=}\,3$ for all main experiments.} Since our method is a plug-in, it can be integrated into existing linear-attention architectures without other modifications. In these experiments, we use FLatten as the baseline linear attention, which combines a focusing function with a lightweight depthwise-convolution branch. In the ablation, we also evaluate additional baselines.

\begin{table}[h]
\vspace{-2mm}
\centering
\setlength{\tabcolsep}{1mm}
\caption{Results of Classification. Comparison of \gls{lsa} across diverse vision backbones. Red indicates gains over the self-attention baseline (in all tables).}
\vspace{-2mm}
\resizebox{\columnwidth}{!}{
\begin{tabular}{l|c|c c|c}
\toprule[1pt]
Model & Input & Params\,(M) & FLOPs\,(G) & Top1-acc(\%) \\
\midrule[0.5pt]

\multicolumn{5}{c}{DeiT Family} \\
\midrule[0.125pt]
DeiT-T~\cite{deit} & 224 & 5.72 & 1.26 & 72.2 \\
FLatten-DeiT-T~\cite{flatten} & 224 & 6.19 & 1.15 & 74.1 \\
\rcol\textbf{\gls{lsa}-DeiT-T} & 224 & 6.30 & 1.29 & \textbf{74.54\improve{2.34}} \\
\midrule[0.125pt]
DeiT-T~\cite{deit} & 384 & 5.79 & 4.70 & 74.9 \\
FLatten-DeiT-T~\cite{flatten} & 384 & 7.14 & 3.37 & 76.53 \\
\rcol\textbf{\gls{lsa}-DeiT-T} & 384 & 7.24 & 3.75 & \textbf{77.54\improve{2.64}} \\
\midrule[0.125pt]
DeiT-S~\cite{deit} & 224 & 22.05 & 4.61 & 79.8 \\
FLatten-DeiT-S~\cite{flatten} & 224 & 22.98 & 4.39 & 80.1 \\
\rcol\textbf{\gls{lsa}-DeiT-S} & 224 & 23.08 & 4.66 & \textbf{80.78\improve{0.98}} \\

\midrule[0.5pt]
\multicolumn{5}{c}{PVT Family} \\
\midrule[0.125pt]
PVT-T~\cite{pvt} & 224 & 13.23 & 1.94 & 75.1 \\
FLatten-PVT-T~\cite{flatten} & 224 & 12.14 & 2.00 & 77.8 \\
\rcol\textbf{\gls{lsa}-PVT-T} & 224 & 12.16 & 2.12 & \textbf{78.53\improve{3.43}} \\
\midrule[0.125pt]
PVT-S~\cite{pvt} & 224 & 24.48 & 3.82 & 79.8 \\
FLatten-PVT-S~\cite{flatten} & 224 & 21.66 & 3.95 & 81.7 \\
\rcol\textbf{\gls{lsa}-PVT-S} & 224 & 21.69 & 4.18 & \textbf{81.96\improve{2.16}} \\

\midrule[0.5pt]
\multicolumn{5}{c}{Swin Family} \\
\midrule[0.125pt]
Swin-T~\cite{SwinTransformer} & 224 & 28.29 & 4.49 & 81.3 \\
FLatten-Swin-T~\cite{flatten} & 224 & 29.71 & 4.49 & 82.1 \\
\rcol\textbf{\gls{lsa}-Swin-T} & 224 & 29.74 & 4.72 & \textbf{82.25\improve{0.95}} \\

\bottomrule[1pt]
\end{tabular}
}
\vspace{-3mm}
\label{tab:classification}
\end{table}


\begin{table}[t]
    \vspace{-2mm}
  \centering
  \caption{Four-stage pyramid architecture: classification and segmentation results. Segmentation is with S\text{-}FPN. Dashes indicate values not reported.}
  \vspace{-2mm}
  \setlength{\tabcolsep}{1.5mm}
  \resizebox{\linewidth}{!}{
  \begin{tabular}{l|c c|c c}
    \toprule[1pt]
    Model & Par.\,(M) & Top1-acc(\%) & mIoU & mAcc \\
    \midrule[0.5pt]
    BiFormer-T~\cite{biformer}                & 13.1  & 81.4 & --- & --- \\
    \rcol\textbf{\gls{lsa}-BiFormer-T}        & 13.2  & \textbf{82.4\improve{1.0}} & --- & --- \\
    BiFormer-S~\cite{biformer}                & 25.5  & 83.8 & 48.9 & --- \\
    \rcol\textbf{\gls{lsa}-BiFormer-S}        & 25.6  & \textbf{84.1\improve{0.3}} & \textbf{49.5\improve{0.6}} & \textbf{61.5} \\
    BiFormer-B~\cite{biformer}                & 56.8  & 84.3 & --- & --- \\
    \rcol\textbf{\gls{lsa}-BiFormer-B}         & 56.8 & \textbf{85.0\improve{0.7}} & --- & --- \\
    \midrule[0.5pt]
  \end{tabular}}
  \label{tab:biformer_cls_then_seg} 
  \vspace{-5mm}
\end{table}

\mypar{Results.} Table~\ref{tab:classification} compares ImageNet Top-1 accuracy across DeiT, PVT, and Swin backbones. With comparable parameter counts and FLOPs, \gls{lsa} consistently outperforms the baseline and FLatten. On DeiT at $224$ input resolution, \gls{lsa} improves DeiT-T/S by $+2.34\%$/$+0.98\%$ and further surpasses FLatten-DeiT-T/S by $+0.44\%$/$+0.68\%$. On PVT, \gls{lsa} yields large gains over the baselines ($+3.43\%$ on PVT-T and $+2.16\%$ on PVT-S) and improves over FLatten by $+0.73\%$/$+0.26\%$. On Swin, \gls{lsa} also brings $+0.95\%$ improvement compare to self-attention. At a higher resolution ($384{\times}384$) on DeiT-T, \gls{lsa} reaches $77.54\%$, improving over FLatten by $+1.01$ points ($76.53\%$), while using fewer FLOPs than the original DeiT-T at the same input size. Table~\ref{tab:biformer_cls_then_seg} shows \gls{lsa} transfers to BiFormer as well, improving classification by $+1.0\%$/$+0.3\%$/$+0.7\%$ for T/S/B.

\subsection{Semantic Segmentation} \label{sec:segmentation}

\mypar{Datasets.} For semantic segmentation we use ADE20K \cite{ade20k}, the standard semantic-segmentation benchmark, which includes 20K training images, 2K validation images, and 150 categories.

\mypar{Settings.} For a fair comparison, we use the same baseline training setup. We plug our method into two representative segmentation frameworks, Semantic FPN \cite{semanticfpn} and UPerNet \cite{upernet}, implemented in MMSegmentation \cite{mmsegmentation}. Backbones are initialized from ImageNet-1K. We train Semantic FPN following the PVT recipe \cite{pvt} and UPerNet using the default Swin configuration \cite{SwinTransformer}.

\begin{table}[h]
    \vspace{-2mm}
    \centering
    \caption{Results of semantic segmentation. FLOPs are computed with an input of $512\times 2048$. S\text{-}FPN denotes SemanticFPN~\cite{semanticfpn}. Dashes indicate values not reported.}
    \vspace{-3mm}
    \resizebox{\linewidth}{!}{%
    \setlength{\tabcolsep}{1mm}
    \begin{tabular}{l|c|c c|c c}
        \toprule[1pt]
        \multicolumn{6}{c}{\textbf{Semantic Segmentation on ADE20K}} \\
        \midrule[0.5pt]
        Backbone & Method & FLOPs\,(G) & Par.\,(M) & mIoU & mAcc \\
        \midrule[0.5pt]
        PVT\text{-}T & S\text{-}FPN & 158 & 17 & 36.57 & 46.72 \\
        FLatten-PVT\text{-}T & S\text{-}FPN & 169 & 16 & 37.21 & 48.95 \\
        \rcol\textbf{\gls{lsa}\text{-}PVT\text{-}T} & S\text{-}FPN & 170 & 16 & \textbf{40.85\improve{4.28}} & \textbf{51.75\improve{5.03}} \\
        \midrule[0.5pt]
        Swin\text{-}T & UperNet & 945 & 60 & 44.51 & 55.61 \\
        FLatten-Swin\text{-}T & UperNet & 946 & 60 & 44.82 & 57.01 \\
        \rcol\textbf{\gls{lsa}\text{-}Swin\text{-}T} & UperNet & 947 & 60 & \textbf{45.03\improve{0.52}} & \textbf{56.4\decrease{0.61}} \\

        \bottomrule[1pt]
    \end{tabular}
    }
    \vspace{-2mm}
    \label{tab:seg}
\end{table}

\mypar{Results.} As shown in Table \ref{tab:seg}, our method delivers consistent gains across all settings. On PVT, it improves mIoU by more than $4.0$ points and mAcc by more than $5.0$ points over the baselines, under comparable FLOPs and parameter counts. On Swin, it improves mIoU of the baseline by $0.5$ points respectively. \gls{lsa}-BiFormer-S (Table~\ref{tab:biformer_cls_then_seg}) also improves the corresponding mIoU baseline by $0.6$ points. These results show that local attention effectively captures informative features for segmentation. 

\subsection{Object Detection} \label{sec:detection}

\mypar{Datasets.} For object detection we use the COCO object detection and instance segmentation benchmark~\cite{coco}, which includes 118K training images and 5K validation images. 

\mypar{Settings.} We assess our method within Mask R-CNN~\cite{mrcn}, initializing backbones with ImageNet pretraining. Experiments use $1 \times$ schedules with varying detection heads.

\mypar{Results.} As shown in Table~\ref{tab:detection}, our method achieves the best accuracy with near-zero compute overhead. On PVT: box AP and mask AP improve by $3.8$ and $2.5$ points, respectively, compared with the PVT baseline. On Swin: box AP and mask AP improve by $0.8$ and $0.5$ points, respectively, compared with the Swin baseline

\begin{table}[h]
\vspace{-2mm}
\centering
\caption{Results of object detection. The FLOPs are computed over backbone, FPN and detection head with input resolution of 1280$\times$800.}
\vspace{-3mm}
\resizebox{\linewidth}{!}{
\setlength{\tabcolsep}{1mm}{
\renewcommand\arraystretch{1.1}
\resizebox{\linewidth}{!}{%
\begin{tabular}{l|c|c|ccc|ccc}
    \toprule
    \multicolumn{9}{c}{\textbf{Mask R-CNN Object Detection on COCO}} \\
    \hline
    Method & FLOPs & Sch. & AP$^b$ & AP$^b_\text{50}$ & AP$^b_\text{75}$ & AP$^m$ & AP$^m_\text{50}$ & AP$^m_\text{75}$ \\
    
    \hline PVT-T 
    & 240G & 1x      & 36.7 & 59.2 & 39.3 & 35.1 & 56.7 & 37.3 \\
    FLatten-PVT-T
    & 244G & 1x      & 38.2 & 61.6 & 41.9 & 37.0 & 57.6 & 39.0 \\
    \rcol \gls{lsa}-PVT-T
    & 245G & 1x      & \bf 40.5\improve{3.8}& \bf 62.9\improve{3.8}& \bf 44.1\improve{4.8}& \bf 37.6\improve{2.5}& \bf 59.7\improve{3.0} & \bf 40.2\improve{2.9}\\

    \hline Swin-T 
    & 267G & 1x      & 43.7 & 66.6 & 47.7 & 39.8 & 63.3 & 42.7 \\
    FLatten-Swin-T
    & 268G & 1x      & 44.2 & 67.3 & 48.5 & 40.2 & 63.8 & 43.0 \\
    \rcol \gls{lsa}-Swin-T
    & 269G & 1x      & \bf 44.5\improve{0.8} & \bf 67.4\improve{0.8} & \bf 48.7\improve{1.0} & \bf 40.3\improve{0.5} & \bf 63.9\improve{0.6} & \bf 43.1\improve{0.4} \\
  
    \bottomrule
\end{tabular}}}
}
\vspace{-2mm}
\label{tab:detection}
\end{table}

\subsection{Ablation Study} 

We conduct ablations to analyze the impact of individual components in our method. \red{Unless stated otherwise, all ablation results are reported on the mini-ImageNet dataset \cite{miniImageNet} ($T=5$) for computational efficiency.}

\mypar{Comparison with other \gls{la} methods.} For a fair comparison with prior \gls{la} modules, we compare against six designs: Hydra Attention~\cite{hydra}, Vanilla \gls{la}~\cite{tranformers_rnns}, Efficient Attention~\cite{efficientattn}, Linear Angular Attention~\cite{castling}, Enhanced \gls{la}~\cite{efficientvit}, and Focused \gls{la} (FLatten)~\cite{flatten}. Table~\ref{tab:other_la} shows that \gls{lsa} outperforms all alternatives, including \gls{soa}.

\begin{table}[h]
    \vspace{-2mm}
    \centering
    \caption{Comparison of \gls{la} variants on DeiT-S (ImageNet).}
    \vspace{-5mm}
    \setlength{\tabcolsep}{1mm}
    \resizebox{.6\linewidth}{!}{
    \begin{tabular}{l|c}
    \toprule[1pt]         
         Linear Attention & Top1-acc(\%)\\
         \midrule[0.125pt]
         DeiT-S~\cite{deit} & 79.8 \\
         Hydra Attn~\cite{hydra} & 75.38 \\
         Vanilla Linear Attn~\cite{tranformers_rnns} & 76.47 \\
         Efficient Attn~\cite{efficientattn} & 76.81 \\
         Linear Angular Attn~\cite{castling} & 79.21 \\
         Enhanced Linear Attn~\cite{efficientvit} & 77.17 \\
         Focused Linear Attn~\cite{flatten} & 80.1 \\
         \rcol\gls{lsa} & \textbf{80.78\improve{0.98}} \\
         \bottomrule[1pt]
    \end{tabular}}
    \label{tab:other_la}
    \vspace{-3mm}
\end{table}

\mypar{Ablation for $\alpha$.} Table~\ref{alpha_ablation} studies $\alpha$, which balances \gls{lsa}’s global and local branches. We compare global-only ($\alpha{=}0$, F{+}D{+}G), local-only ($\alpha{=}1$, F{+}D{+}L), and learned $\alpha$. On both DeiT tiny and small, learning $\alpha$ performs best ($76.3\%$ and $80.7\%$), outperforming both individual branches. Moreover, global-only exceeds local-only ($73.7$ vs.\ $73.4$ on DeiT-T; $78.3$ vs.\ $78.1$ on DeiT-S), suggesting that long-range transitive aggregation provides a stronger standalone signal than local neighborhood matching.

\begin{table}[h]
\vspace{-2mm}
\caption{Component ablation on DeiT-T/DeiT-S classification.}
\vspace{-2mm}
\centering
\footnotesize
\renewcommand{\arraystretch}{1.05}
\setlength{\tabcolsep}{5pt}
\resizebox{\linewidth}{!}{
\begin{tabular}{c|cc>{\columncolor{Dandelion!12}}c}
\toprule
Backbone & F{+}D{+}G ($\alpha{=}0$) & F{+}D{+}L ($\alpha{=}1$) & \bf Learned $\alpha$ \\
\midrule
DeiT-T & 73.7 & 73.4 & \textbf{76.3} \improve{2.6} \\
DeiT-S & 78.3 & 78.1 & \textbf{80.7} \improve{2.4} \\
\bottomrule
\end{tabular}
}
\label{alpha_ablation}
\vspace{-4mm}
\end{table}

\begin{figure}[h]
\vspace{-5pt}
\centering
\resizebox{0.72\linewidth}{!}{%
\begin{tikzpicture}[x=1.2cm, y=1cm]

\foreach \y in {0,1,2,3,4,5}{
    \draw[gray!20] (0.7,\y) -- (5.3,\y);
}
\foreach \x in {1,2,3,4,5}{
    \draw[gray!20] (\x,0) -- (\x,5);
}

\draw[thick, ->, >=stealth] (0.7,0) -- (5.5,0);
\draw[thick, ->, >=stealth] (0.7,0) -- (0.7,5.3);

\foreach \y/\v in {
    0/72,1/73,2/74,3/75,4/76,5/77
}{
    \node[left] at (0.65,\y) {\small \v};
}

\foreach \x in {1,...,5}{
    \node[below] at (\x,-0.05) {\small \x};
}

\draw[blue!75!black, thick]
    (1,1.34) --
    (2,3.43) --
    (3,3.63) --
    (4,3.16) --
    (5,3.73);

\foreach \x/\y in {
    1/1.34, 2/3.43, 3/3.63,
    4/3.16, 5/3.73
}{
    \fill[blue!75!black] (\x,\y) circle (2.5pt);
}

\node[right=4pt, font=\scriptsize] at (1,1.34)
    {$73.34 \pm 0.64$};

\node[above left=2pt, font=\scriptsize] at (2.3,3.43)
    {$75.43 \pm 0.56$};

\node[above, font=\scriptsize] at (3,3.63)
    {$75.63 \pm 0.39$};

\node[below, font=\scriptsize] at (4,3.16)
    {$75.16 \pm 0.13$};

\node[above, font=\scriptsize] at (5,3.73)
    {$\mathbf{75.73 \pm 0.47}$};

\node at (3,-0.75) {Truncation depth ($T$)};
\node[rotate=90] at (-0.15,2.5)
    {Top-1 accuracy (\%)};

\end{tikzpicture}%
}
\vspace{-4pt}
\caption{Effect of truncation length $T$ on DeiT-T
accuracy, reported as mean $\pm$ standard deviation
across three runs.}
\label{fig:t_seed_variability}
\vspace{-7pt}
\end{figure}
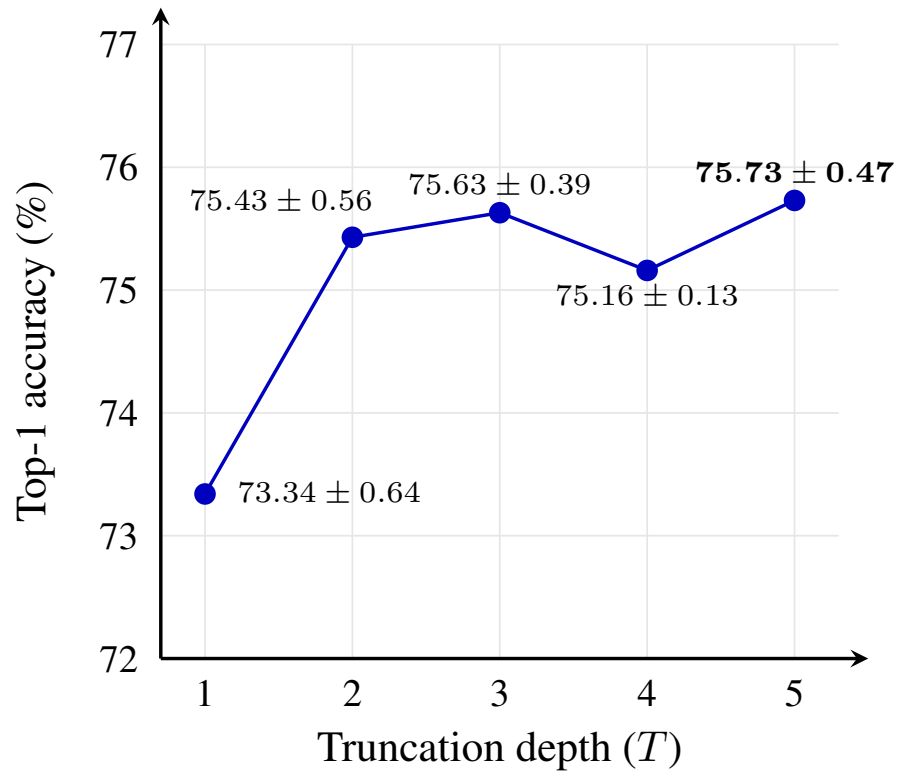

\mypar{Number of truncations.} We vary the truncation length $T$ in Eq.~\ref{eq:geometric}. All LiSA components and settings are kept fixed; only the truncation length $T$ is varied. Figure~\ref{fig:t_seed_variability} shows that accuracy peaks at a moderate truncation length ($T=5$). \red{Across three seeds, $T>1$ consistently outperforms $T=1$, isolating the benefit of higher-order propagation. Among the higher-order settings, $T=3$ achieves the second-highest mean accuracy ($75.63 \pm 0.39$), closely approaching the best-performing $T=5$ ($75.73 \pm 0.47$). We use $T=5$ for the ablation studies due to its highest mean accuracy, while adopting $T=3$ for the main ImageNet experiments to reduce computational cost by requiring fewer propagation terms with minimal loss in accuracy.}

\mypar{Statistical variability.} \red{In Figure~\ref{fig:t_seed_variability}, we report results across three runs for truncation depths $T=1$ to $T=5$, including the mean and standard deviation. The results confirm that higher-order propagation consistently improves over $T=1$ (e.g., $75.63 \pm 0.39$ for $T=3$ and $75.73 \pm 0.47$ for $T=5$, versus $73.34 \pm 0.64$ for $T=1$), with $T=5$ achieving the highest mean accuracy. The relatively small differences among the stronger higher-order settings, together with their consistent gains over $T=1$, indicate that the benefit is robust to run-to-run variability rather than driven by a single run.}

\mypar{Source of improvement.} GL is our proposed attention core, a direct plug-in replacement for LA. F and D are additions borrowed from FLatten and are applied identically on top of both LA and GL. Our contribution is to replace the vanilla linear attention core (LA) with a higher-order global-local stochastic diffusion operator (GL). Table~\ref{component} isolates the contribution of our core: First, \textbf{GL $>$ LA} (69.2 vs.\ 65.7), showing that our core already improves over vanilla linear attention before introducing F or D. Second, the same trend holds when each auxiliary component is added independently: \textbf{GL+F $>$ LA+F} (69.3 vs.\ 65.9) and \textbf{GL+D $>$ LA+D} (75.2 vs.\ 73.3). Notably, \textbf{GL+D} even surpasses the full FLatten baseline \textbf{LA+F+D} (75.2 vs.\ 74.4). Finally, when both F and D are applied equally to both cores, \textbf{GL+F+D $>$ LA+F+D} (76.3 vs.\ 74.4), showing that LiSA outperforms FLatten under the same auxiliary design. These consistent gains indicate that the improvement comes from our proposed GL mechanism rather than from the borrowed auxiliary components.

\begin{table}[h]
\vspace{-2mm}
\centering
\caption{Component ablation on DeiT-T classification.}
\vspace{-2mm}
\label{component}
\footnotesize
\renewcommand{\arraystretch}{1.05}
\setlength{\tabcolsep}{5pt}
\resizebox{\linewidth}{!}{
\begin{tabular}{cc|cc|cc|c>{\columncolor{Dandelion!12}}c}
\toprule
LA & GL & LA{+}F & GL{+}F & LA{+}D & GL{+}D & \shortstack{\textbf{FLatten}\\LA{+}F{+}D} & \shortstack{\textbf{LiSA}\\GL{+}F{+}D} \\
\midrule
65.7 & 69.2 & 65.9 & 69.3 & 73.3 & 75.2 & 74.4 & \textbf{76.3} \\
\bottomrule
\end{tabular}
}
\vspace{-4mm}
\end{table}

\mypar{Different \gls{la} baseline.} This experiment highlights the plug-in nature of our method and evaluates whether \gls{lsa} generalizes beyond FLatten. On DeiT-T, with linear angular attention~\cite{castling}, the baseline reaches $66.0\%$ Top-1, while adding \gls{lsa} improves it to $73.4\%$. Furthermore, integrating \gls{lsa} into Agent Attention~\cite{agent_attention} improves DeiT-T from 73.40 to 75.77 (+2.37) and DeiT-S from 78.83 to 80.84 (+2.01), demonstrating that \gls{lsa} transfers effectively to other \gls{la} variants.

\begin{figure}[t]
    \centering

    \begin{subfigure}[t]{\linewidth}
        \centering
        \includegraphics[width=\columnwidth]{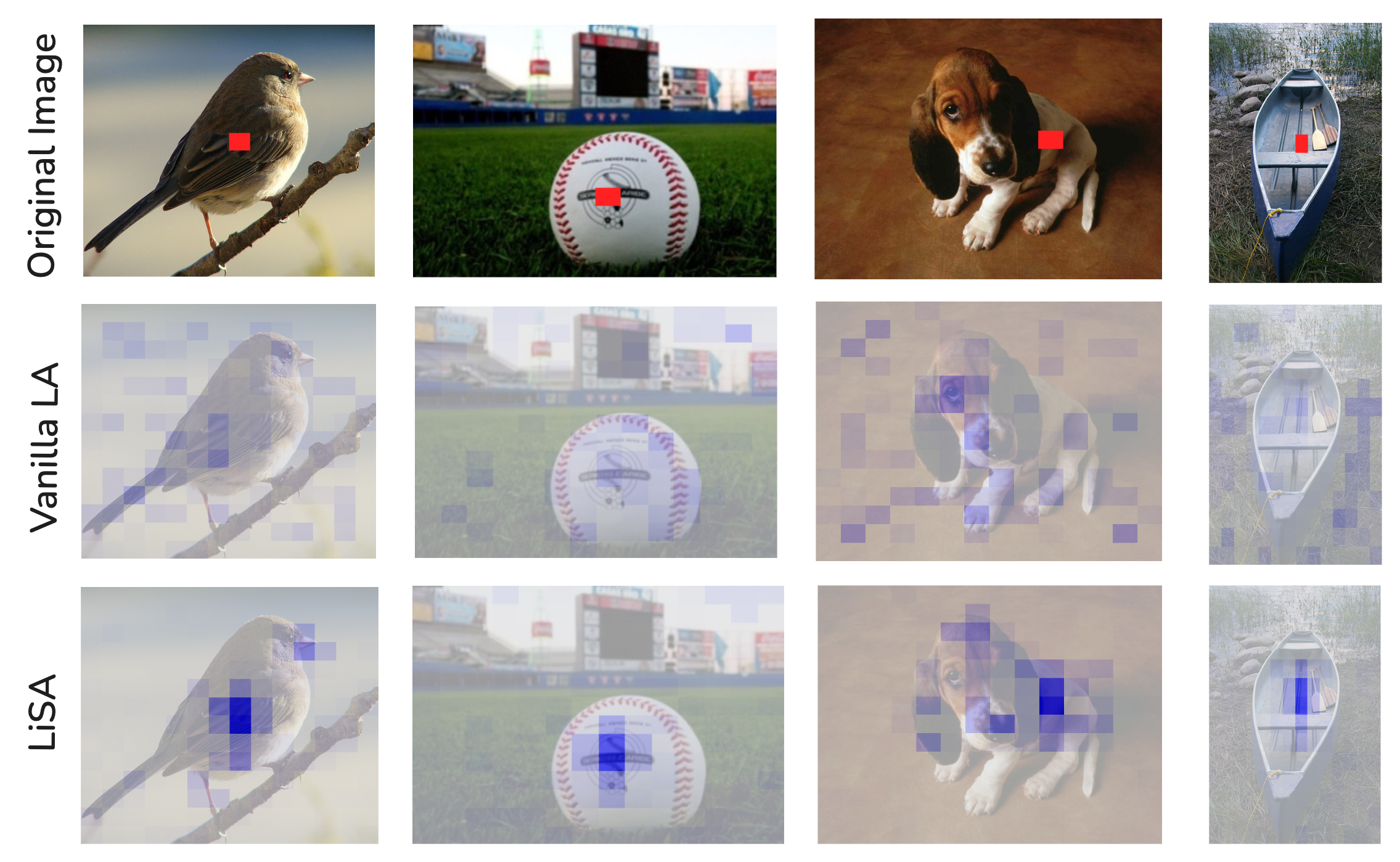}
        \caption{DeiT-T attention map. The red square marks the query patch.}
        \label{fig:attn}
    \end{subfigure}


    \begin{subfigure}[t]{\linewidth}
        \centering
        \includegraphics[width=\columnwidth]{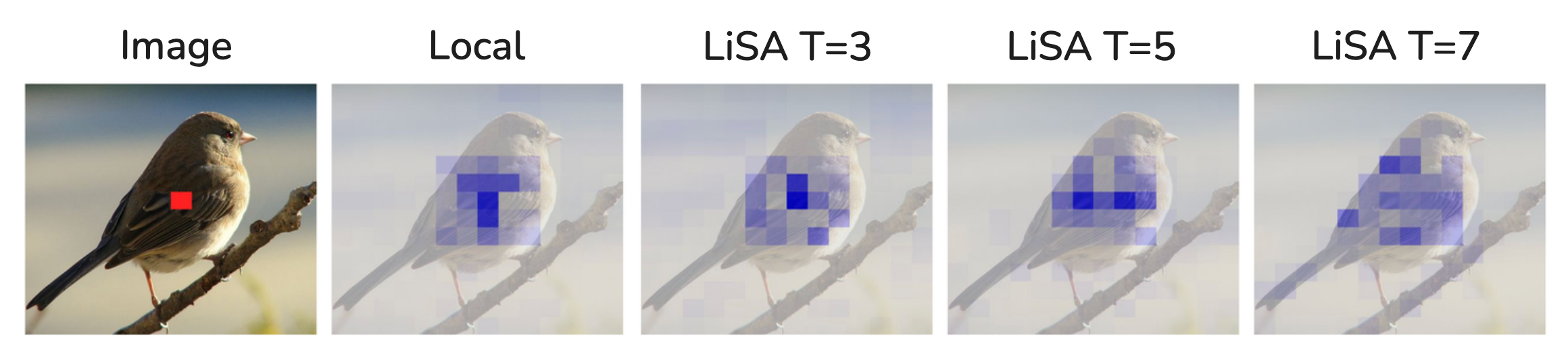}
        \caption{Transitive reasoning visualized in DeiT-T attention maps.}
        \label{fig:transitive}
    \end{subfigure}

    \caption{Visualization of attention behavior in DeiT-T.}
    \label{fig:attention_visualization}
    \vspace{-5mm}
\end{figure}

\mypar{Distribution of \gls{la}.} As shown in Figure~\ref{fig:attn}, vanilla \gls{la} produces diffuse maps: the weights are spread across many locations, so each query effectively aggregates an almost averaged representation of the whole image. In contrast, \gls{lsa} yields more structured, non-uniform patterns, where larger weights are assigned to patches on the same object as the query, while most background areas receive only small contributions.

\mypar{Visualization of Transitive Reasoning.} In Figure~\ref{fig:transitive}, we start from a query patch on the bird (red box). Local attention remains restricted to nearby patches, so its influence stays spatially confined. In contrast, \gls{lsa} expands its receptive field with hop depth: as we increase the truncation length ($T{=}3,5,7$), attention progressively propagates along semantically related regions, reaching distant yet correlated parts of the bird and eventually covering the tail and most of the object. Please zoom in for a clearer view.

\mypar{Practical efficiency.} We provide latency and peak GPU memory plots across sequence lengths. The efficiency advantage of LiSA over softmax attention becomes more pronounced as sequence length increases. DeiT-T scales quadratically in both latency and memory, becoming prohibitively expensive at long sequences, while FLatten and LiSA maintain linear scaling. LiSA incurs a modest latency overhead compared to FLatten, which is expected given its higher-order random-walk modeling and adaptive local branch. This overhead reflects a trade-off between efficiency and expressiveness. Memory-wise, LiSA preserves linear scaling and remains substantially more efficient than softmax attention, with some additional overhead over FLatten due to its higher-order modeling.

\begin{figure}[h]
\vspace{-8pt}
\centering
\begin{subfigure}{0.49\columnwidth}
    \centering
    \includegraphics[width=\linewidth]{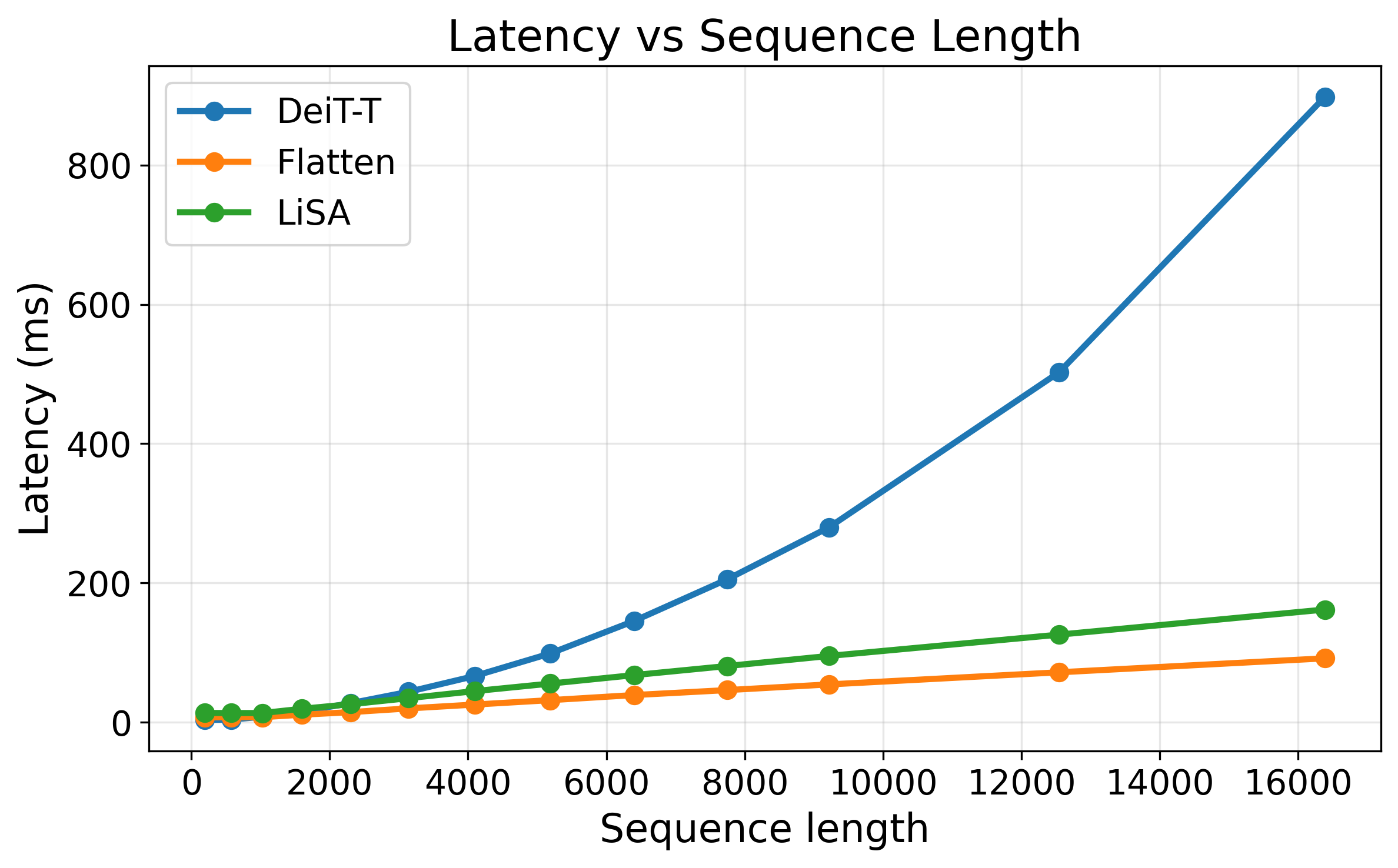}
    \caption{Latency}
    \label{fig:latency_seq}
\end{subfigure}
\hfill
\begin{subfigure}{0.49\columnwidth}
    \centering
    \includegraphics[width=\linewidth]{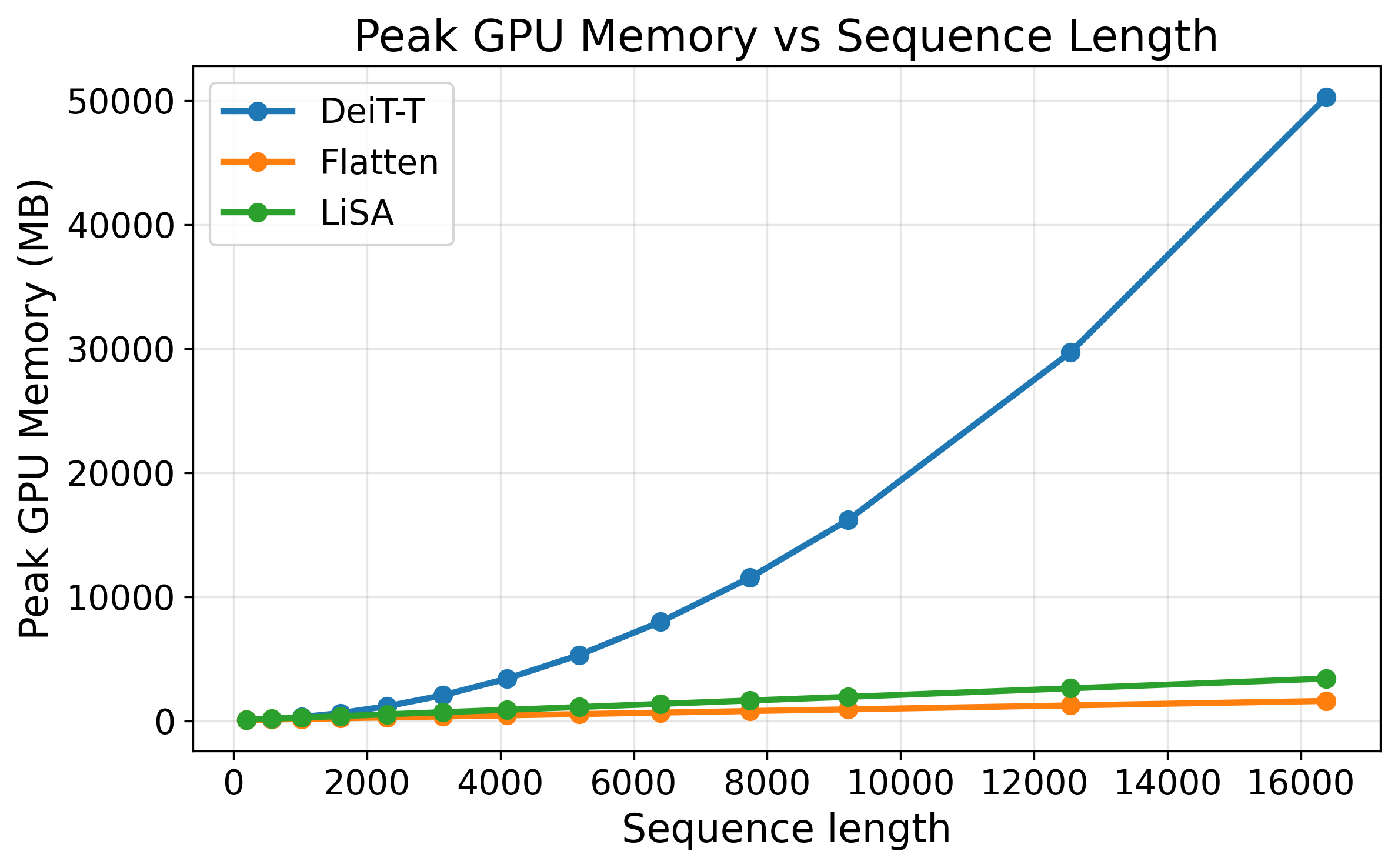}
    \caption{GPU memory}
    \label{fig:memory_seq}
\end{subfigure}
\vspace{-7pt}
\caption{Efficiency comparison across sequence lengths.}
\label{fig:latency_memory_seq}
\vspace{-12pt}
\end{figure}

\vspace{-2pt}
\section{Conclusion}
We introduced \gls{lsa}, a \gls{la} plug-in module that restores higher-order token interactions via a truncated random-walk formulation, while preserving stability and locality through an adaptive $3{\times}3$ neighborhood. This design directly overcomes the first-order limitation of vanilla \gls{la}, enabling richer transitive reasoning without sacrificing linear complexity. Across image classification, semantic segmentation, and object detection, \gls{lsa} yields consistent gains at near-identical compute budgets and transfers cleanly to diverse backbones (DeiT, PVT, Swin, BiFormer). Ablations confirm that each component contributes meaningfully and that a short truncation length already offers a strong accuracy–latency trade-off, positioning \gls{lsa} as a more powerful drop-in alternative to standard kernelized \gls{la}. As future work, we plan to integrate \gls{lsa} into additional state-of-the-art \gls{la} variants and study its behavior under diverse training recipes and data regimes.

\newpage
{
    \small
    \bibliographystyle{ieeenat_fullname}
    \bibliography{main}
}
\clearpage
\setcounter{page}{1}
\maketitlesupplementary

\section{Proofs of the Propositions}
This section provides proofs for the propositions introduced in the methodology.

\begin{proposition}
For walks of infinite length, probabilities can be computed analytically as
\begin{equation}\label{eq:inverse}
P^{(\infty)} \, = \, (1\!-\!\beta)\big(I - \beta S\big)^{-1}
\end{equation}
\end{proposition}

\begin{proof}
Consider the sum over the first $T$ terms in Eq. (4). Pre-multiplying this sum by $(I-\beta S)$ gives
\red{\begin{align}
(I-\beta S) P^{(T)} & \, = \, \left(\frac{1\!-\!\beta}{1\!-\!\beta^{T}}\right)\left(\sum_{k=0}^{T-1} \beta^{k} S^{k} - \sum_{k=1}^{T} \beta^{k} S^{k}\right)\nonumber\\
& \, = \,  \left(\frac{1\!-\!\beta}{1\!-\!\beta^{T}}\right)\left(I - (\beta S)^{T}\right).
\end{align}}
Since $\beta\!<\!1$, we have that \red{$(\beta S)^{T} \!\to\! 0$ and $\beta^{T} \!\to\! 0$ when $T \!\to\! \infty$, hence}
\begin{align}
(I-\beta S) & P^{(\infty)} \, = \, (1\!-\!\beta) \nonumber\\
& P^{(\infty)} \, = \, (1\!-\!\beta)\big(I - \beta S\big)^{-1}.
\end{align}
\end{proof}

\begin{proposition}
\label{theorem2}
\red{Let $R_T$ be the truncation residual obtained after retaining the first $T$ terms in $P^{(\infty)}$. The $L_1$ norm of $R_T$ decays exponentially as the truncation depth $T$ increases.}
\red{\begin{equation}
R_T \, = \, (1\!-\!\beta)\left(\sum_{k=T}^{\infty}\!\!\beta^k S^k\right). 
\end{equation}}
\end{proposition}

\begin{proof}
\red{We use the entrywise matrix $L_1$ norm, defined as
$\|A\|_1 = \sum_{i,j} |A_{ij}|$. Since $S$ is stochastic, we have that $S^k$ is also stochastic, hence $\|S^k\|_1 = n$. Therefore, 
\begin{align}
\|R_T\|_1
&= (1\!-\!\beta)\sum_{k=T}^{\infty}\beta^k \|S^k\|_1 \nonumber\\
&= n(1\!-\!\beta)\sum_{k=T}^{\infty}\beta^k \nonumber\\
&= n(1\!-\!\beta)
\left(
\sum_{k=0}^{\infty}\beta^k
-
\sum_{k=0}^{T-1}\beta^k
\right) \nonumber\\
&= n(1\!-\!\beta)
\left(
\frac{1}{1\!-\!\beta}
-
\frac{1\!-\!\beta^{T}}{1\!-\!\beta}
\right) \nonumber\\
&= n\beta^{T}.
\end{align}
For $\beta<1$, $\|R_T\|_1$ therefore decays exponentially to zero as the truncation length $T$ increases.}
\end{proof}

\section{Additional Ablation Study}
\red{Unless stated otherwise, all ablation results are reported on the mini-ImageNet dataset \cite{miniImageNet} ($T=5$) for computational efficiency.}

\mypar{Impact of kernel functions.} \red{In Table~\ref{tab:kernel_funcs}, we ablate the feature-map non-linearity used in \gls{la}. ReLU achieves the best performance ($76.3\%$ Top-1), slightly outperforming ELU ($76.0\%$). We therefore use ReLU by default in all experiments.}

\begin{table}[!htbp]
\vspace{-2mm}
  \centering
  \small
  \caption{Impact of different kernel functions (DeiT-T).}
  \vspace{-3mm}
  \setlength{\tabcolsep}{5pt}
    \resizebox{.55\linewidth}{!}{
    \begin{tabular}{l|c>{\columncolor{Dandelion!12}}c}
    \toprule
    Kernel Function & ELU  & ReLU \\
    \midrule
    Accuracy (\%) & 76.0 & \textbf{76.3} \\
    \bottomrule
  \end{tabular}
  }
  \vspace{-3mm}
\label{tab:kernel_funcs}
\end{table}

\mypar{Ablation for local window size.} In Table~\ref{tab:window_size} we vary the window size in local attention on DeiT-T. A compact $3\,{\times}\,3$ window achieves the best Top-1 ($76.3\%$) while being less computationally expensive, whereas larger windows slightly reduce accuracy ($75.7\%$ at $5\,{\times}\,5$, $76.1\%$ at $7\,{\times}\,7$) and increase cost.

\begin{table}[!htbp]
\vspace{-1mm}
\centering
\caption{Ablation study on local window size using DeiT-T.}
\vspace{-2mm}
\setlength{\tabcolsep}{5pt}
\resizebox{.7\linewidth}{!}{
\begin{tabular}{c|>{\columncolor{Dandelion!12}}ccc}
\toprule
Local Window Size & $3 \times 3$ & $5 \times 5$ & $7 \times 7$ \\
\midrule
Accuracy (\%) & \textbf{76.3} & 75.7 & 76.1 \\
\bottomrule
\end{tabular}
}
\label{tab:window_size}
\vspace{-4mm}
\end{table}

\section{Learnable Step Weights Details}

\red{The step weights in Eq.~(10) are predicted per token and per head, allowing each token to adaptively weight the retained diffusion terms through a lightweight gate applied to its query and key:
\begin{equation}
\beta_i^{(j)}
=
\mathrm{softmax}\!\left(
W_2\,\mathrm{ReLU}\!\left(
W_1\left[q_i^{(j)};k_i^{(j)}\right]
\right)\right)
\in \mathbb{R}^{T},
\end{equation}
where $W_1 \in \mathbb{R}^{d_h \times 2d_h}$ and
$W_2 \in \mathbb{R}^{T \times d_h}$. The softmax guarantees
$\beta_k \geq 0$ and $\sum_{k=0}^{T-1}\beta_k = 1$, so the output is a convex combination of the truncation length $T$. The gate parameters are shared across heads and instantiated independently in each layer, while the weights are predicted separately for each token and head. The gate adds $\mathcal{O}(n\,d\,d_h + n\,T\,d)$ operations, preserving linear complexity in the number of tokens and leaving the asymptotic complexity of Algorithm~1 unchanged.}

\section{Failure Cases}

\red{Excessive higher-order diffusion can be detrimental due to over-smoothing: since $S$ is row-stochastic, deep walks converge toward a query-independent stationary distribution, making token representations increasingly similar and weakening discriminative local structure.}

\section{Effective Regimes}

\red{LiSA provides the largest gains precisely where long-range modeling is hardest: plain columnar backbones with weak first-order cores ($+3.43$ on PVT-T; $+7.4$ when added to linear angular attention), high-resolution inputs ($+2.64$ at $384^2$), and dense prediction tasks that require spatial reasoning ($+4.28$ mIoU on ADE20K, $+3.8$ box AP on COCO). Gains remain consistent but naturally smaller for backbones that already encode hierarchy and locality (e.g.\ $+0.95$ on Swin-T), where part of the long-range signal is available by design. Its additional cost is therefore least justified at short sequence lengths, where the $\mathcal{O}(T\,d_h^{3})$ term is not amortized.}

\section{Model Architecture}
We summarize the architectures of the three Transformer models used in the main paper, DeiT, PVT, and Swin Transformer, in Table~\ref{tab:model_deit}, \ref{tab:model_pvt1}, and \ref{tab:model_swin}. In all cases, we replace each original self-attention block with our proposed LiSA block.

\begin{table}[!htbp]
    \caption{Architectures of \gls{lsa}-DeiT models.
    All original DeiT transformer blocks are removed
    (\#DeiT blocks = 0) and replaced with \gls{lsa} blocks.}
    \centering
    \scriptsize
    \setlength{\tabcolsep}{2mm}
    \renewcommand\arraystretch{1.3}
    \resizebox{\linewidth}{!}{%
    \begin{tabular}{l|c|c|c|c}
    \toprule
    Model & Stage & Resolution &
    \shortstack[c]{\gls{lsa} blocks\\
    (res, dim, heads, \#layers)} &
    \#DeiT blocks \\
    \midrule
    \gls{lsa}-DeiT-T & Encoder & $14\times14$ &
    $(14\times14,\ 192,\ 3)\times12$ & 0 \\
    \midrule
    \gls{lsa}-DeiT-S & Encoder & $14\times14$ &
    $(14\times14,\ 384,\ 6)\times12$ & 0 \\
    \bottomrule
    \end{tabular}%
    }
    \label{tab:model_deit}
\end{table}


\begin{table}[!htbp]
    \caption{Architectures of \gls{lsa}-PVT models.
    At all stages, the original PVT transformer blocks are
    removed (\#PVT blocks = 0) and replaced with
    \gls{lsa} blocks.}
    \centering
    \scriptsize
    \setlength{\tabcolsep}{2mm}
    \renewcommand\arraystretch{1.3}
    \resizebox{\linewidth}{!}{%
    \begin{tabular}{l|c|c|c|c}
    \toprule
    Model & Stage & Output &
    \shortstack[c]{\gls{lsa} blocks\\
    (res, dim, heads, \#layers)} &
    \#PVT blocks \\
    \midrule
    \multirow{4}{*}{\gls{lsa}-PVT-T}
    & res1 & $56\times56$ &
    $(56\times56,\ 64,\ 2)\times2$ & 0 \\
    & res2 & $28\times28$ &
    $(28\times28,\ 128,\ 4)\times2$ & 0 \\
    & res3 & $14\times14$ &
    $(14\times14,\ 320,\ 10)\times2$ & 0 \\
    & res4 & $7\times7$ &
    $(7\times7,\ 512,\ 16)\times2$ & 0 \\
    \midrule
    \multirow{4}{*}{\gls{lsa}-PVT-S}
    & res1 & $56\times56$ &
    $(56\times56,\ 64,\ 2)\times3$ & 0 \\
    & res2 & $28\times28$ &
    $(28\times28,\ 128,\ 4)\times4$ & 0 \\
    & res3 & $14\times14$ &
    $(14\times14,\ 320,\ 10)\times6$ & 0 \\
    & res4 & $7\times7$ &
    $(7\times7,\ 512,\ 16)\times3$ & 0 \\
    \bottomrule
    \end{tabular}%
    }
    \label{tab:model_pvt1}
\end{table}

\begin{table}[!htbp]
    \caption{Architecture of the \gls{lsa}-Swin model. At all stages, the original Swin transformer blocks are removed (\#Swin blocks = 0) and replaced with \gls{lsa} blocks.}

    \centering
    \scriptsize
    \setlength{\tabcolsep}{3mm}
    \renewcommand\arraystretch{1.3}
    \resizebox{\linewidth}{!}{%
    \begin{tabular}{c|c|cc}
    \toprule
    \multirow{2}{*}{Stage} & \multirow{2}{*}{Output} &
    \multicolumn{2}{c}{\gls{lsa}-Swin-T} \\
    \cline{3-4}
    & &
    \shortstack[c]{\rule{0pt}{10pt}\gls{lsa} blocks\\(win, dim, heads, \#layers)} & \#Swin blocks \\
    \midrule
    res1 & $56\times56$ &
    $(7\times7,\ 96,\ 3)\times 2$ & 0 \\
    \hline
    res2 & $28\times28$ &
    $(7\times7,\ 192,\ 6)\times 2$ & 0 \\
    \hline
    res3 & $14\times14$ &
    $(7\times7,\ 384,\ 12)\times 6$ & 0 \\
    \hline
    res4 & $7\times7$ &
    $(7\times7,\ 768,\ 24)\times 2$ & 0 \\
    \bottomrule
    \end{tabular}%
    }
    \label{tab:model_swin}
\end{table}

\section{Detailed Efficiency Measurements}

\red{To complement the efficiency plots in the main paper, we additionally report the exact latency, throughput, and peak GPU memory values in Table~\ref{tab:details_efficiency}. The measurements span input resolutions from $224^2$ to $2048^2$, corresponding to sequence lengths from $197$ to $16,385$ tokens, and compare DeiT-T, FLatten, and LiSA under the same setting. Consistent with Fig.~5, DeiT-T exhibits rapidly increasing latency and memory as the sequence length grows, whereas FLatten and LiSA scale much more favorably. LiSA incurs additional latency over FLatten due to its higher-order propagation and adaptive local branch, while retaining substantially lower latency and memory than softmax attention at long sequences.}

\begin{table*}[!t]
\centering
\caption{\red{Latency, throughput, and peak GPU memory across input resolutions.}}
\label{tab:details_efficiency}
\small
\setlength{\tabcolsep}{2.5pt}

\begin{tabular}{cc|ccc|ccc|ccc}
\toprule

\multicolumn{2}{c|}{\textbf{Input}} &
\multicolumn{3}{c|}{\cellcolor{blue!13}\textbf{DeiT-T}} &
\multicolumn{3}{c|}{\cellcolor{gray!20}\textbf{Flatten}} &
\multicolumn{3}{c}{\cellcolor{teal!18}\textbf{LiSA}} \\

\midrule

\textbf{Res.} &
\textbf{Seq.} &

\cellcolor{blue!7}\makecell{\textbf{Latency}\\\textbf{(ms)}} &
\cellcolor{blue!7}\makecell{\textbf{Throughput}\\\textbf{(img/s)}} &
\cellcolor{blue!7}\makecell{\textbf{GPU Memory}\\\textbf{(MB)}} &

\cellcolor{gray!12}\makecell{\textbf{Latency}\\\textbf{(ms)}} &
\cellcolor{gray!12}\makecell{\textbf{Throughput}\\\textbf{(img/s)}} &
\cellcolor{gray!12}\makecell{\textbf{GPU Memory}\\\textbf{(MB)}} &

\cellcolor{teal!10}\makecell{\textbf{Latency}\\\textbf{(ms)}} &
\cellcolor{teal!10}\makecell{\textbf{Throughput}\\\textbf{(img/s)}} &
\cellcolor{teal!10}\makecell{\textbf{GPU Memory}\\\textbf{(MB)}} \\

\midrule

224  & 197   & 3.50   & 2288.00 & 74.0    & 7.15  & 1118.91 & 73.2   & 13.53  & 591.49 & 95.4 \\
384  & 577   & 4.07   & 1967.60 & 152.7   & 7.34  & 1090.32 & 110.2  & 13.52  & 591.60 & 172.1 \\
512  & 1025  & 8.06   & 993.06  & 312.9   & 7.40  & 1080.62 & 151.1  & 13.48  & 593.68 & 263.4 \\
640  & 1601  & 15.56  & 514.19  & 627.9   & 11.11 & 720.42  & 206.6  & 19.47  & 410.90 & 382.0 \\
768  & 2305  & 27.69  & 288.93  & 1178.7  & 14.74 & 542.73  & 275.7  & 26.15  & 305.90 & 529.2 \\
896  & 3137  & 43.80  & 182.66  & 2061.4  & 19.96 & 400.87  & 352.4  & 34.86  & 229.49 & 697.4 \\
1024 & 4097  & 66.51  & 120.29  & 3394.2  & 25.59 & 312.58  & 441.9  & 44.89  & 178.21 & 890.9 \\
1152 & 5185  & 98.78  & 80.99   & 5317.0  & 31.79 & 251.65  & 546.2  & 55.58  & 143.95 & 1115.9 \\
1280 & 6401  & 146.13 & 54.74   & 7976.9  & 39.05 & 204.88  & 663.9  & 67.58  & 118.38 & 1367.2 \\
1408 & 7745  & 205.52 & 38.93   & 11545.5 & 46.43 & 172.29  & 791.3  & 80.73  & 99.10  & 1642.5 \\
1536 & 9217  & 279.54 & 28.62   & 16210.5 & 54.77 & 146.07  & 926.4  & 95.24  & 83.99  & 1937.2 \\
1792 & 12545 & 502.82 & 15.91   & 29690.0 & 71.92 & 111.24  & 1244.8 & 126.33 & 63.33  & 2623.8 \\
2048 & 16385 & 898.92 & 8.90    & 50281.7 & 91.84 & 87.11   & 1605.2 & 161.71 & 49.47  & 3409.1 \\

\bottomrule
\end{tabular}
\end{table*}

\end{document}